\documentclass[a4paper, 10pt, conference]{ieeeconf}      
\usepackage{FG2021}
\usepackage{amssymb}
\usepackage{pifont}
\newcommand{\cmark}{\ding{51}}%
\newcommand{\xmark}{\ding{55}}%
\usepackage{comment}
\usepackage{graphicx}
\usepackage{algorithm2e}
\usepackage{algorithmic}

\FGfinalcopy 

\IEEEoverridecommandlockouts                              
\def\FGPaperID{122} 

\title{\LARGE \bf
Detecting Human-to-Human-or-Object (\hho{}) Interactions with \dia{}
}

\author{\parbox{16cm}{\centering
    {\large Astrid Orcesi$^{1,2}$ \hspace{0.5cm} Romaric Audigier$^{1,2}$ \hspace{0.5cm} Fritz Poka Toukam$^1$ \hspace{0.5cm} Bertrand Luvison$^{1,2}$}\\
    {\normalsize
    $^1$ Université Paris-Saclay, CEA, List, F-91120, Palaiseau, France\\
    $^2$ Vision Lab, ThereSIS, Thales SIX GTS, Campus Polytechnique, Palaiseau, France}}\\
    {\tt\small \{firstname.lastname\}@cea.fr}
}

\newcommand{\ao}[1]{\textcolor[rgb]{0,0,0}{#1}}
\newcommand{\ra}[1]{\textcolor[rgb]{0,0,0}{#1}}
\newcommand{\bl}[1]{\textcolor[rgb]{0,0,0}{#1}}
\usepackage{array}
\newcolumntype{L}[1]{>{\raggedright\let\newline\\\arraybackslash\hspace{0pt}}m{#1}}
\newcolumntype{C}[1]{>{\centering\let\newline\\\arraybackslash\hspace{0pt}}m{#1}}
\newcolumntype{R}[1]{>{\raggedleft\let\newline\\\arraybackslash\hspace{0pt}}m{#1}}

\newcommand{\eg}[0]{\textit{e.g.}}

\def\dia{{DIABOLO}}
\def\hho{{$H^2O$}}
\def\hhoi{{$H^2OI$}}
\def\hoi{{$HOI$}}
\def\hhi{{$HHI$}}

\begin{document}

%
%
%




\IEEEoverridecommandlockouts\pubid{\makebox[\columnwidth]{978-1-6654-3176-7/21/\$31.00~\copyright{}2021 IEEE \hfill}
\hspace{\columnsep}\makebox[\columnwidth]{ }}

\ifFGfinal
\thispagestyle{empty}
\pagestyle{empty}
\else
\author{Anonymous FG2021 submission\\ Paper ID \FGPaperID \\}
\pagestyle{plain}
\fi
\maketitle

\begin{abstract}
Detecting human interactions is crucial for human behavior analysis. Many methods have been proposed to deal with Human-to-Object Interaction (\hoi{}) detection, i.e., detecting in an image which person and object interact together and classifying the type of interaction. However, Human-to-Human Interactions, such as social and violent interactions, are generally not considered in available \hoi{} training datasets. As we think these types of interactions cannot be ignored and decorrelated from \hoi{} when analyzing human behavior, we propose a new interaction dataset to deal with both types of human interactions: Human-to-Human-or-Object (\hho{}). In addition, we introduce a novel taxonomy of verbs, intended to be closer to a description of human body attitude in relation to the surrounding targets of interaction, and more independent of the environment. Unlike some existing datasets, we strive to avoid defining synonymous verbs when their use highly depends on the target type or requires a high level of semantic interpretation. As \hho{} dataset includes V-COCO images annotated with this new taxonomy, images obviously contain more interactions. This can be an issue for \hoi{} detection methods whose complexity depends on the number of people, targets or interactions. Thus, we propose DIABOLO (Detecting InterActions By Only Looking Once), an efficient subject-centric single-shot method to detect all interactions in one forward pass, with constant inference time independent of image content. In addition, this multi-task network simultaneously detects all people and objects. We show how sharing a network for these tasks does not only save computation resource but also improves performance collaboratively. Finally, DIABOLO is a strong baseline for the new proposed challenge of \hho{}-Interaction detection, as it outperforms all state-of-the-art methods when trained and evaluated on \hoi{} dataset V-COCO. We hope that this new dataset and new baseline will foster future research.
\ao{\hho{} is available on https://kalisteo.cea.fr/.}
\end{abstract}

\section{INTRODUCTION}

One of the requirements for visual analysis of human behavior is to recognize human actions.
Many methods~\cite{carreira2017quo, andriluka20142d, soomro2012ucf101, sigurdsson2016hollywood} have dealt with human action recognition in video. The goal is to classify a whole video clip containing a single main action. However, substantial video datasets containing multiple simultaneous actions with localization of their subjects and targets are still missing.
Other methods~\cite{gkioxari2018detecting, gao2018ican, calipso} have dealt with the so-called Human-Object Interaction (\hoi{}) detection task from a single image. It consists in determining and locating the list of triplets $<subject, verb, target>$ which describe all the simultaneous interactions in an image.
Several image datasets for \hoi{} have been made available~\cite{vcoco, hicodet}.
However, they focus on targets which are non-human, called ``objects'' hereafter.
Therefore, a lot of human interactions, such as social or violent interactions, are not taken into consideration. 
To analyze human behavior, Human-to-Human Interactions (\hhi{}), i.e. interactions between people, cannot be ignored and decorrelated from \hoi{}.
For example, in video surveillance applications, it is interesting to recognize fighting people, to distinguish them from hugging people, and also to detect people kicking urban equipment.
The lack of datasets dealing with both types of human interactions is the first motivation for proposing a new dataset called \hho{} (Human-to-Human-or-Object).

Second, verb taxonomies used by existing datasets~\cite{vcoco, hicodet} are sometimes ambiguous.
For example, some types of interactions correspond to synonymous verbs (\eg, to inspect or to watch an object, to hold or to carry a cell phone, to read or to look at a book...) which sometimes only differ from the type of target (\eg, to surf, to snowboard, or to skateboard) or requires a high level of semantic interpretation of the context or the intent (\eg, to hold, to pick or to buy an apple, to ride or to sit on a horse). Conversely, some English verbs can merge different types of human postures or attitude that could be distinguished in another language (\eg, to ride a horse or a bus do not correspond exactly to the same body attitude in relation to the target object).
This motivates us to introduce a novel taxonomy of verbs, intended to be closer to a description of the human body attitude in relation to the surrounding targets of interaction, and less dependent on the environment, the target type or the linguistic arbitrariness. 
To constitute \hho{} dataset, we re-annotated images from V-COCO dataset~\cite{vcoco} with this new taxonomy  including both object and human targets, and added new images to enrich the dataset with \hhi{} verbs.

Consequently, this new dataset motivated us to propose a novel method named DIABOLO (Detecting InterActions By Only Looking Once) to address \hho{} Interaction (\hhoi{}) detection problem.
Formally, image-based \hoi{} or \hhoi{} detection can be decomposed in three steps: detecting objects and people (named together ``instances'' hereafter) in the image, pairing interacting instances, and classifying the interaction. 
Most state-of-the-art approaches~\cite{ulutan2020vsgnet, wan2019pose, sun2020human}, considered as two-stage methods, rely on an external object detector to perform the first step (identify candidates) then feed a second network with the detections to perform pairing and classification.
Other methods~\cite{kim2020uniondet, dirv} include all steps in the same multi-task network. However, they dissociate the learning of the object detection task from the interaction detection task. 
We propose to study the effect of learning both tasks jointly or consecutively and how these tasks can collaborate to improve overall performance while reducing the computation resource with the use of a shared backbone.

Finally, for many applications such as video surveillance, ambient assisted living or cobotics, the response time of the method is an important criterion in order not to induce a latency in the real-time analysis. 
Moreover, the image may have a lot of people and objects that possibly interact together. Ideally, computation time of the method should not be affected by the density of the image. 
However, most of state-of-the-art methods cannot guarantee a constant time, as an interaction estimation network is applied on all possible pairs. 
These methods are not scalable with the number of visible instances and interactions.
Instead, the so-called single-shot methods use a single forward pass of the image in the network, which ensures constant computation time.

The contributions of this paper are the following:
\begin{itemize}
\item We propose a new dataset \hho{} to train and evaluate models on the novel task of Human-to-Human-or-Object Interaction (\hhoi{}) detection, not limited to non-human targets.
\item We introduce a new taxonomy of verbs intended to be closer to the human body attitude relative to the surrounding targets of interaction, and to be less dependent on the context or environment in which the interactions occur. We strove to avoid synonyms or linguistic bias and to reduce room for high-level interpretation.
\item We propose \dia{}, a single-shot method which detects instances of the image and estimates their interactions in a single forward pass throughout the image. This new subject-centric method runs in fast and constant time independently of the image content, and without needing external object detector.
\item \dia{} is a first yet strong baseline for this new \hhoi{} challenge. Indeed, when trained and evaluated on a \hoi{} detection challenge like V-COCO, \dia{} outperforms all existing state-of-the-art methods.
\item Finally, ablative experiments on \dia{} show how multi-task learning can improve performance of interaction detection.
\end{itemize}

The paper is organized as follows:
Related work is introduced in Section~\ref{sec:sota}. Sections~\ref{sec:h2o} and \ref{sec:diabolo} present \hho{} dataset and \dia{} method, respectively. Evaluations of \dia{} on both \hoi{} and \hhoi{} detection challenges are presented in Section~\ref{sec:exp}, as well as an ablative study and a comparison with state-of-the-art methods.

\section{RELATED WORK}
\label{sec:sota}

\subsection{HOI datasets}

Human behavior understanding has often resulted in action recognition in video clip. A significant number of datasets~\cite{carreira2017quo, andriluka20142d, soomro2012ucf101, sigurdsson2016hollywood} consist of video clips of few seconds 
which have to be classified among a set of possible actions. 
The large dataset of video clips AVA \cite{gu2018ava} partially introduces spatial and temporal localization of the actions: only person bounding boxes with related actions are annotated in the central frame of each clip. No information is provided about the targets of the actions.

More recently, some image datasets focus on relationship between objects in an image. For instance, Visual Genome \cite{krishna2017visual} proposes 108,000 images annotated with 18 visual relationships. 
The relationships can involve any two objects of the image. But predicates have rather low semantic levels and mostly concern positional relationships (as ``behind'' or ``next to''). This dataset is usually used to study visual relationship, but not human interactions. HCVRD~\cite{zhuang2018hcvrd} is a subset of Visual Genome which selects human-centered relationships expanded into 927 categories, including actions, (pre)positional and comparative relations.
This large number of interactions involves that most of them appear less than 10 times.
HICO~\cite{chao2015hico} is a dataset for \hoi{} classification: the images are subject-centric and the whole image corresponds to a list of non-localized interactions. HICO-DET~\cite{hicodet} uses 47,776 images from HICO and adds bounding boxes annotations to create an \hoi{} detection dataset. HICO-DET contains 117 verbs some of which are very specific to a target object category or synonyms. Finally, V-COCO \cite{vcoco} is a subset of 10,346 images from COCO \cite{coco} annotated with 26 interaction verbs, but the interaction target is limited to a single object per verb.

\ao{Regarding \hhi{} datasets, TVHI \cite{hoai2014talking} proposes only few social interactions. \cite{choi2012unified} focuses only on relative motion between people. None of these datasets propose to merge \hoi{} and \hhi{}.} Therefore, we define a new taxonomy of 51 verbs (cf. Section~\ref{taxonomy}) including both \hoi{} and \hhi{} and avoiding synonyms and target specificities, to be less dependent on the environment and closer to the human body attitude relative to the surrounding targets of interaction. Then we propose the \hho{} dataset that includes V-COCO images re-annotated with this taxonomy and 3,666 new images collected to enrich \hhi{}. 

\subsection{HOI detection methods}

\hoi{} detection in images consists in localizing interacting instances in the scene and pairing them with a specific interaction verb to create a list of triplets $<subject, verb, target>$ of all the visible interactions.

On the one hand, most of the previous methods adopt a two-stage strategy \cite{gkioxari2018detecting, gao2018ican, calipso, li2019transferable, zhou2020cascaded, ulutan2020vsgnet, wan2019pose, liu2020consnet, sun2020human}. During the first stage, they use an external object detector as \cite{girshick2015fast} to point out interacting candidates, and then, during the second stage, another network is dedicated to estimate interactions between the proposals. First works are essentially based on the appearance of the objects: \cite{gkioxari2018detecting, gao2018ican, ulutan2020vsgnet} extract features from the object location to classify the interactions. More recently, improvement in the second stage have been proposed by using additional information in the image. For instance, \cite{li2019transferable, wan2019pose, liu2020consnet} use human pose to have a finer analysis of the posture of the interaction subject. Other methods add word embedding \cite{sun2020human, lu2016visual} or segmentation \cite{zhou2020cascaded}. Finally, \cite{sun2020human} combines all these kinds of additional information. The major drawback of these methods is that they analyze each possible $<subject, object>$ pair in order to determine all the interactions in the image. Their computation time is therefore quadratic with the number of instances in the image.
Some methods provide an alternative to studying all possible pairs and thus accelerate the inference time.
\cite{liao2020ppdm, wang2020learning} model the interaction detection problem as a keypoint detection one. \cite{calipso} densely estimate interactions over an anchor grid to have computation time independent of the number of instances in the image, then use an external object detector to point out the anchors that actually correspond to instances. 

On the other hand, some recent works propose one-stage \hoi{} detectors. \cite{kim2020uniondet, dirv} only rely on regression and classification to predict interactions. They both integrate an instance detection branch similar to classical object detectors. Interaction detection is then based on the union of regressed bounding boxes~\cite{kim2020uniondet}. \cite{dirv} notes that it is better to focus at regions of interaction rather than the whole union box which presents too much unnecessary information. Therefore, \cite{dirv} proposes an interaction region-centric branch to detect interaction. These two methods initialize their internal object detector with a pre-trained model and then freeze these weights to learn the interaction branch.

Our method is one-stage and contrary to \cite{dirv}, DIABOLO is subject-centric and uses embedding to pair interacting instances. 
Unlike \cite{sun2020human, lu2016visual, zhou2020cascaded}, we do not use any additional information. 
Finally, DIABOLO is the first multi-task network which trains object and interaction detections jointly. 

\section{PROPOSED DATASET}
\label{sec:h2o}

In this section, we present \hho{} Dataset, an image dataset annotated for Human-to-Human-or-Object interaction detection.
We first present the modalities to constitute the dataset, the taxonomy chosen to annotate interactions and compare it to currently available datasets. Then, we present metrics to evaluate algorithm performance.

\subsection{\hho{} composition}

\hho{} is composed of the 10,301 images from V-COCO \cite{vcoco} images to which are added 3,666 images  \ao{selected in the wild as for COCO dataset \cite{coco} and which} mostly contain interactions between people. Thus, unlike current available datasets, \hho{} presents interactions between human and object but also human and human. 
As for object annotations, all interacting instances are annotated with bounding boxes, even if they do not belong to the 80 classes of COCO \cite{coco}. In total, instances are distributed in 214 classes. However, out of a total of 128,969 annotated instances (58,225 people and 70,744 objects), 96\% are part of the 80 classes of COCO.
Interactions are exhaustively annotated for each person, whether they are with an object or another person.
\ao{These annotations were made with Pixano (https://pixano.cea.fr) annotation tool.}

\subsection{\hho{} taxonomy}
\label{taxonomy}

\begin{figure}[h!]
\centering
\includegraphics[width=8.5cm]{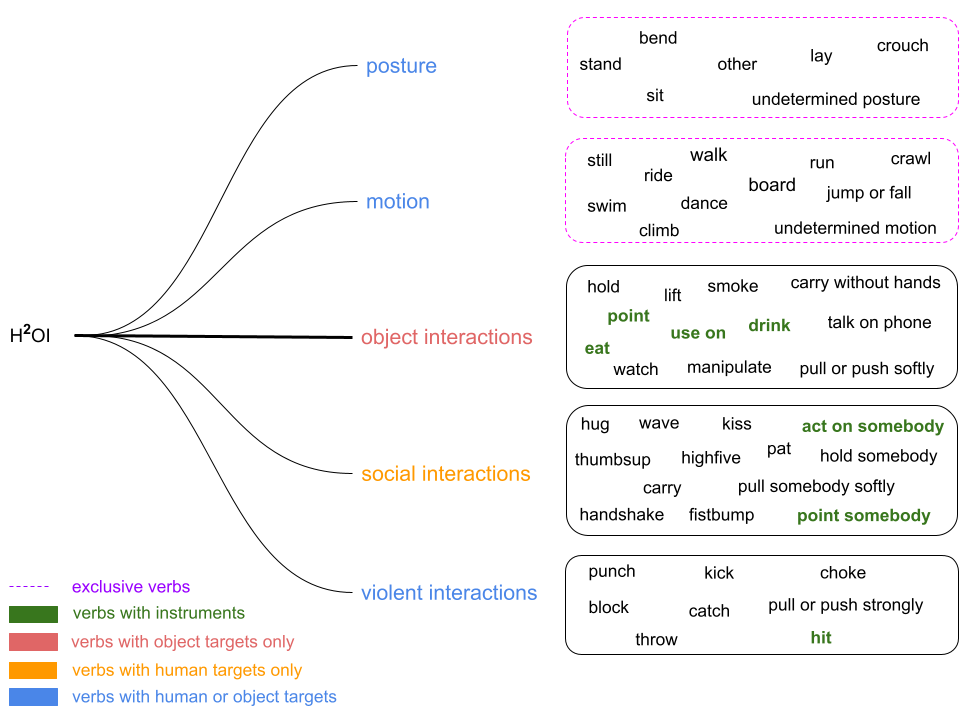}
\caption{\hho{} taxonomy hierarchy}
\label{h2o_taxonomy}
\end{figure}

To annotate \hho{}, we defined a new taxonomy of verbs including both \hoi{} and \hhi{}. 
This taxonomy (cf. Figure~\ref{h2o_taxonomy}) is intended to be closer to the human body attitude relative to the surrounding targets of interaction, and less dependent on the environment in which the interactions occur, the target type or the linguistic bias.
So we strive to avoid synonymous verbs when their use highly depends on the target type or verbs which require a high level of semantic interpretation.

\hho{} dataset is annotated with 51 verbs divided into five categories: (i) verbs describing the general posture of the subject, (ii) verbs related to the way the subject is moving, (iii) verbs used for interactions with objects, (iv) verbs describing human-to-human interactions and finally (v) verbs of interactions involving strength or violence which can affect either objects or people.

Posture and motion categories have verbs which are exclusive and mandatory. This means that each person in the image must be annotated with a single verb of posture and a single verb of motion. 
\hho{} contains 58,225 posture verbs and as many motion verbs.
Verbs of the other three categories are not exclusive, nor mandatory. This means that subjects can be annotated with none, one or several of these interaction verbs.

\textit{Posture verbs --}
General posture verbs are: ``stand'', ``bend'', ``sit'', ``crouch'', ``lay'', ``other'' and ``undetermined posture''. ``Undetermined posture'' is dedicated to truncated people whose posture cannot be determined for sure.
On the contrary, ``other'' means the subject is fully seen but his/her posture is not usual or cannot be simply described.
For example, it is the case of some acrobatic positions in sports images.

\textit{Motion verbs --} 
Motion verbs are: ``still'', ``walk'', ``run'', ``ride'', ``board'', ``crawl'', ``jump or fall'', ``dance'', ``swim'', ``climb'' and ``undetermined motion''. ``Still'' is dedicated to people who do not move. ``Undetermined motion'' is annotated for people who are truncated and whose movement cannot be described for sure. We chose to annotate with ``board'' all types of movement on a board (\eg, skateboard, surf, snowboard, skis) not to be tied to the context of the interactions.
All posture and motion verbs can have a target and it is necessarily the same for both verbs. For instance, a person can be ``stand''-ing ``still'' on a stool. In the third row of Figure~\ref{h2o_vs_vcoco}, the distinction between posture and motion with the possibility of annotating a target allow a precise description of a ``crouching man boarding a skateboard''. 

\textit{Interactions with object --} These verbs can be done only with objects. We finely separate the fact of holding something in four verbs: ``hold'', ``lift'', ``carry without hands'' and ``pull or push softly'' as they are visually different. ``Lift'' is dedicated to heavy objects which need two hands to be lifted (\eg, a sofa). ``Carry without hands'' is used with objects which are carried without the need of hands (\eg, handbag or backpack on the back or  over the shoulder). ``Pull or push softly'' is reserved for rolling objects as suitcase, shopping cart or stroller. We do not distinguish ``pull'' from ``push'' as it is ambiguous on only one image. ``Manipulate'' is annotated for subjects who use an object for its specific function. For example, this verb gathers verbs as ``cut'', ``brush'' or ``stick''. Four verbs of this category accept until two types of interacting objects: the final target of the interaction and the tool or instrument used to execute the interaction on the target object. These verbs are: ``point'', ``use on'', ``eat'' and ``drink''. ``Eat'' and ``drink'' are annotated as such, only when the subject makes the gesture of bringing something to the mouth (\eg, to sit around a table with a dish does not mean the person is eating it). Finally, ``watch'', ``talk on phone'' and ``smoke'' are also part of this category.

\textit{Social interactions --} Verbs of this category are exclusively related to interactions between people. The interaction verbs are: ``hug'', ``kiss'', ``handshake'', ``wave'', ``highfive'', ``fistbump'', ``thumbsup'', ``pat'', ``hold somebody'', ``pull or push somebody softly'', ``carry somebody'', ``point somebody'' and ``act on somebody''. ``Point'' and ``act on'' can be used with the instrument (if any) that allows the interaction achievement, as well as the final human target (\eg, a doctor ``acts on'' a patient ``with'' a stethoscope).

\textit{Violent interactions --} Targets of the interaction verbs of this category can be either a person or an object. Interactions are performed with strength or violence. The verbs selected highly depend on the body parts involved: ``punch'', ``kick'', ``choke'', ``block'', ``pull or push strongly'', ``throw'', ``catch'' and ``hit''. If appropriate, ``hit'' can be annotated with an instrument (\eg, ``hit'' the ball ``with'' a baseball bat).

\subsection{Comparison with existing datasets}
\label{DatasetComparison}

\begin{figure}[h!]
\centering
\includegraphics[width=7.5cm]{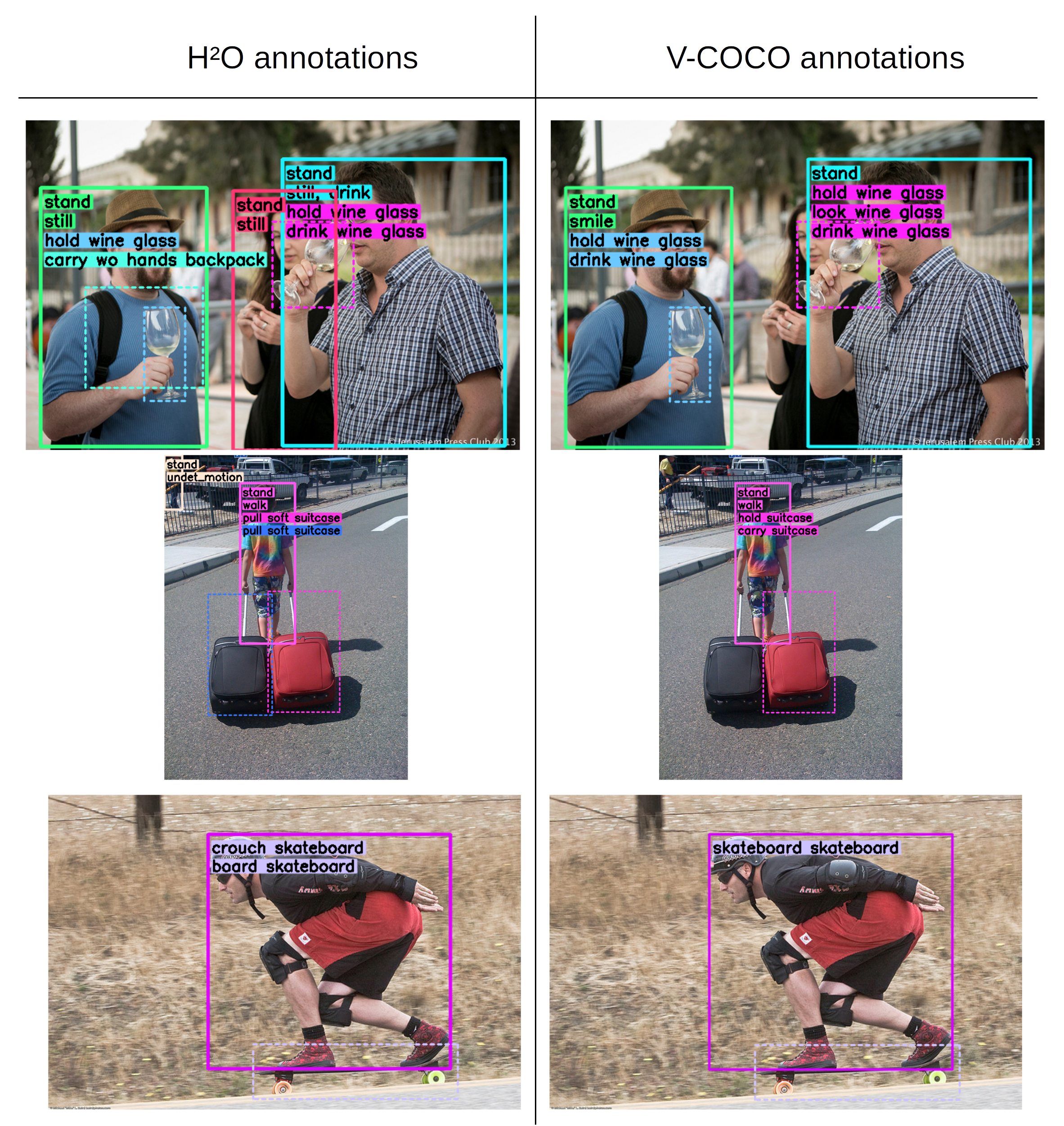}
\caption{Comparison between \hho{} and V-COCO annotations}
\label{h2o_vs_vcoco}
\end{figure}

V-COCO dataset \cite{vcoco} is a subset of COCO dataset \cite{coco} where each person is annotated with 26 interaction verbs over 80 object categories. Contrary to \hho{}, four verbs (``stand'', ``smile'', ``run'' and ``walk'') do not allow target. In addition, each interaction in restricted to only one target object for a given subject. These two points do not allow to describe the scene exhaustively. For example, if a person is standing on a stool and holding two different objects at the same time, these limitations force the partial description of a standing person holding one single object. In the image on second row of Figure~\ref{h2o_vs_vcoco}, V-COCO annotates only one of the two suitcases pulled by the person. Moreover and contrary to \hho{}, not all interactions that are part of the 26 verbs are exhaustively annotated in the image (\eg, the standing still woman in the first row of Figure~\ref{h2o_vs_vcoco} is not annotated). Figure~\ref{h2o_vs_vcoco} illustrates differences between \hho{} and V-COCO annotations.

HICO-DET dataset \cite{hicodet} is larger and more diverse than V-COCO dataset because it contains 117 action categories over the same 80 object categories as COCO dataset. However, these predicates are very specific and too much linked to the context of the scene. For instance, ``dribble'' is always done with a sports ball or ``grind'' is always done with a board.

Unlike \hho{}, in these two datasets, interacting objects outside the 80 classes of COCO are not annotated.
Table~\ref{datasets} and \ref{datasets2} present statistics on \hho{} compared to V-COCO and HICO-DET. Table~\ref{datasets} shows the amount of images, the number of verbs and their target type. Table~\ref{datasets2} compares the overall number of interactions between \hho{} and V-COCO, and details the numbers of interactions by categories. As posture and motion categories are mandatory, therefore exhaustively annotated in \hho{}, their quantity is much larger than for V-COCO. The table also compares the mean number of people per image, and the mean number of objects per image.

\begin{table}[h!]
\caption{Dataset comparison according to number of images and verbs}
\label{datasets}
\begin{center}
\begin{tabular}{|l|c|c|c|}
\hline
\textbf{Dataset} & \textbf{\#images} & \textbf{\#verbs} & \textbf{Target type}\\
\hline
\hline
HICO-DET \cite{hicodet} & 47,774 & 117 & object\\
V-COCO \cite{vcoco} & 10,346 & 26 & object\\
\hline
\hho{} & 13,967 & 51 & object, person\\
\hline
\end{tabular}
\end{center}
\end{table}

\begin{table}[h!]
\caption{Dataset comparison according to the amount of annotations}
\label{datasets2}
\begin{center}
\begin{tabular}{|c|lr|c|c|}
\hline
\textbf{Dataset} & \multicolumn{2}{c|}{\textbf{\#interactions}} & \textbf{\#person} & \textbf{\#object}\\
& & & \textbf{per} & \textbf{per}\\
& & & \textbf{image} & \textbf{image}\\
\hline
\hline
V-COCO \cite{vcoco} & \textbf{Total:} & \textbf{49,019} & 3.8 & 4.9\\
& Posture + Motion: & 22,480  & & \\
& \hoi{}: & 26,539 & & \\
\hline
\hho{} & \textbf{Total:} & \textbf{151,816} & 4.2 & 5.1\\
& Posture + Motion: & 116,450& & \\
& \hoi{}: & 25,984 & & \\
& Social: & 5,413 & & \\
& Violence: & 3,969 & & \\
\hline
\end{tabular}
\end{center}
\end{table}

\subsection{Evaluation protocols}
\label{EvalH2O}

V-COCO \cite{vcoco} proposes an evaluation of the \hoi{} detections based on two metrics. The first one is the agent mean average precision, $AP_{agent}$ which measures the accuracy of the pair $<subject, verb>$. The second one, the role mean average precision called $AP_{role}$ analyzes the whole triplet $<subject, verb, target>$ considering as a true positive when all components are correct.
The predicted human and object bounding boxes are supposed to be correct if they have an IoU greater than 0.5 with ground truth boxes. 
Two different $AP_{role}$ metrics are proposed. They differ in the evaluation of specific interaction triplets $<subject, verb, \emptyset>$ that appear when the target object is either not seen, not existing or not in the 80 classes of COCO. In the first one ($AP_{role1}$), not predicting the $\emptyset$ as target is penalized whereas in the second scenario ($AP_{role2}$), it is not. Moreover, V-COCO dataset assumes only a single target object for a given verb and a given person. Consequently, $AP_{role}$ computation is limited to this assumption.


\bl{\hho{} dataset provides annotation of targets even though target object is not part of the 80 classes of COCO. The specific triplet $<subject, verb, \emptyset>$ in \hho{} means target is not seen or not existing. Consequently, we propose a new scenario called ``Objectness" where target object outside the 80 classes of COCO should be detected and properly associated to the interaction. Class label of such a target is  ``other''.}

\bl{\hho{} also provides several objects in interaction for a given verb and a given person if they exist. Consequently, $AP_{role}$ metric has been adapted to take into account this new feature. For clarity, original V-COCO scenario will be called ``Original" as opposed to the new ``Objectness" scenario proposed in this paper. Both dataset and evaluation code will be made available.}

\begin{figure*}[h!]
\centering
\includegraphics[width=15.5cm]{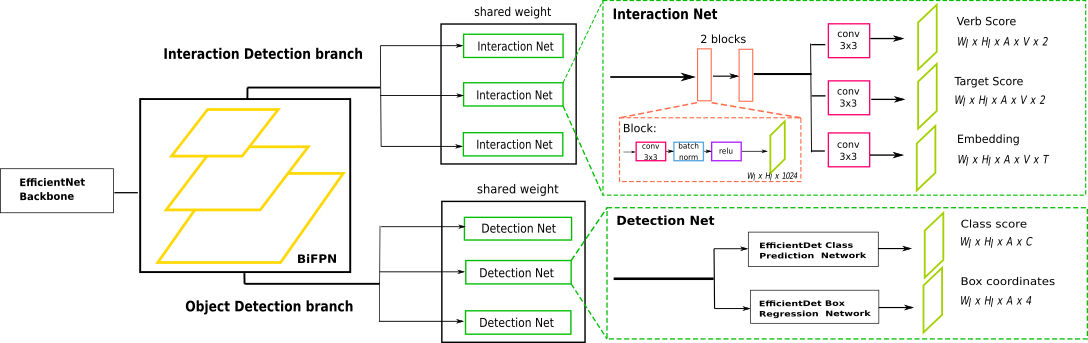}
\caption{DIABOLO architecture network with two branches for interaction detection (top) and object detection (bottom). $W_{l}$ and $H_{l}$ denote respectively the width and the height of the feature map of the pyramid at level $l$. $A$ is the number of anchors, $V$ is the number of verbs and $T$ is the embedding size.}
\label{diabolo}
\end{figure*}

\section{PROPOSED METHOD}
\label{sec:diabolo}

In this section, we present our approach for interaction detection. The method is named DIABOLO and is based on the CALIPSO \cite{calipso} method. CALIPSO densely estimates interactions on \ao{a classical detection} anchor grid thanks to three tasks. The first task defines the action score of each anchor. The second task estimates for each verb the presence of an interacting target for each human anchor. The third task gives an embedding for each anchor in the image in order to associate the interacting pairs. Thus, at the inference time, CALIPSO needs an external object detector to point out anchors which actually correspond to people or objects. Contrary to CALIPSO, DIABOLO integrates an object detector, based on EfficientDet \cite{efficientdet}. Therefore DIABOLO simultaneously detects people, objects and their interactions in the image in a single shot.

Figure \ref{diabolo} illustrates the multi-task neural network architecture used by DIABOLO. The object detection and the interaction estimation branches share the same EfficientNet backbone followed by a BiFPN introduced by \cite{efficientdet}.

\subsection{Object detection branch}

DIABOLO uses EfficientDet \cite{efficientdet} as object detector. EfficientDet is an anchor-based architecture as most of object detector. The outputs of the object detection branch are the classification of the instances and the regression of their bounding boxes. During training, DIABOLO supervises these two tasks in the same way as \cite{efficientdet}. 

\subsection{Interaction estimation branch}
\label{interactbranch}

DIABOLO interaction estimation branch is close to the CALIPSO one. Indeed, this branch densely estimates three tasks on the anchor grid:  verb prediction task \bl{in active and passive voice}, target presence estimation task and interaction embedding task. The use of an EfficientNet backbone and BiFPN instead of a ResNet FPN allows to reduce the number of convolutions contained in the interaction network. Unlike CALIPSO, DIABOLO interaction network is composed of only two convolutional blocks after the BiFPN. The supervision of the target presence estimation task and the interaction embedding task are the same as in CALIPSO \ao{i.e. respectively a binary cross entropy loss and a metric learning pull-push loss.}
However, DIABOLO uses a focal loss \cite{focalloss} to densely predict the interaction verb.
The focal loss $FL$ aims to better cope with the class occurrence imbalance and is computed as follows:
\begin{equation}
FL = \sum_{v \in V} \sum_{a \in A^+} -\alpha(1-p_a^v)^\gamma \log (p_a^v) \\
\end{equation}
where $v$ is a verb in the set $V$, $\alpha$ and $\gamma$ are focal loss parameters and $A^+$ denotes the set of anchors associated to an interacting object. \ao{In focal loss, $p$ being the model's estimated probability for class $v$ and anchor $a$, $p_a^v$ is equal to $p$ if the ground truth class of anchor $a$ is $v$, $1-p$ otherwise.}


\subsection{Inference}

From the detection branch, people and objects are pointed out with a traditional Non Maxima Suppression (NMS) algorithm 
except that indices of selected anchors are used to directly read the information relative to the different interactions in the interaction branch. 
For each detected person, an interaction score for each verb and each possible target for this verb, according to its category (cf. Figure.\ref{h2o_taxonomy}), is computed similarly to \cite{calipso}. Categories of exclusive interactions are processed independently by providing only triplets of the verb with the highest probability in its category.


\section{EXPERIMENTS}
\label{sec:exp}

In this section, we present the experiments to evaluate DIABOLO performance relative to the state of the art and propose a first baseline on \hho{} dataset.

\subsection{Datasets and Metrics}

\subsubsection{Datasets}
We chose to evaluate DIABOLO on V-COCO \cite{vcoco} and on our new dataset \hho{}. 
As mentioned in section \ref{DatasetComparison}, V-COCO still has the drawback of defining predicates too closely related to the context, and HICO-DET has the same issue to a greater extent.

\subsubsection{Evaluation metrics}

To evaluate results on V-COCO \cite{vcoco}, we use their standard evaluation setting, as presented in section~\ref{EvalH2O}, using $AP_{role1}$ scenario which is the most difficult one. Notice that similarly to previous work~\cite{gkioxari2018detecting, gao2018ican}, the verb ``point'' is not taken into account since it has too few samples.
On \hho{}, both ``Original" and ``Objectness" scenarios are used for evaluation.

\subsection{Implementation details}

The EfficientNet backbone, the BiFPN and the detection branch are initialized with weights previously learned on the object detection COCO dataset \cite{coco}. We use EfficientDet-D3 for fair comparison with the state of the art and EfficientDet-D1 for ablation studies as it is lighter than D3.
For multi-task learning, we compare the strategies of learning instance detection branch (on COCO) and freezing it or not during the subsequent learning of interaction detection, with the strategy of continuing learning instance detection along with interaction detection.
In the latter strategy, we use V-COCO or \hho{} to train the whole network of DIABOLO and, at the same time, COCO (V-COCO images excluded) to continue training the detection branch only.
We will see that V-COCO data are not varied enough to correctly learn instance detection.
Indeed, V-COCO has only 5,400 training images whereas EfficientDet is usually learned on the 120,000 training images of COCO. 
That is why we use mixed batches with images from V-COCO and images from COCO to improve the object detection branch. 
Proportions of images from each dataset in the batch are constant during training (batch of 24 COCO + 24 \hho{} images on \hhoi{} challenge and see Table~\ref{multi_dataset} for batch composition on V-COCO \hoi{} challenge).
%
Two different trainings are made according to evaluation scenarios mentioned in Section~\ref{EvalH2O}. For the ``Original'' one, only target objects within 80 classes of COCO are taken into account and for the ``Objectness'' one, all unusual objects excluded by the 80 classes are added in a new class labeled as ``other'' that detection branch has to learn.
Training is performed on NVIDIA A100-SXM4 GPUs. DIABOLO is trained with stochastic gradient descent (SGD), with an initial learning rate of 0.016, which is then reduced by 10 at 10,000 iterations. Focal loss parameters $\alpha$ and $\gamma$ are respectively set to $0.25$ and $2$. Horizontal image flipping and color jittering are applied for data augmentation.

\subsection{Results of DIABOLO on V-COCO and comparison with \hoi{} state of the art}

Table~\ref{results_vcoco} presents the results of DIABOLO on V-COCO dataset compared to \ao{the best} state-of-the-art methods \ao{and to one-stage methods.} 
DIABOLO is the top-1 method outperforming current state-of-the-art method DIRV \cite{dirv} by 12.1 p.p with 76.7\% for $AP_{agent}$ and 1.2 p.p with 57.3\% for $AP_{role}$. We believe that such an increase in $AP_{agent}$ can be explained by the subject-centric aspect of DIABOLO that is more suitable for multiple action recognition. Notice that $AP_{role}$ result for DIABOLO is obtained without using any ontology information, contrary to \cite{dirv} which loses 1.3 p.p. if no ontology information is provided.

\begin{table}[h!]
\caption{State-of-the-art methods for \hoi{} detection: Performances evaluated on V-COCO test set}
\label{results_vcoco}
\begin{center}
\begin{tabular}{|L{2.3cm}|c|c|c|}
\hline
\textbf{Method} & \textbf{One Stage} & \textbf{$AP_{agent} (\%)$} & \textbf{$AP_{role} (\%)$} \\
\hline
\hline
InteractNet \cite{gkioxari2018detecting} & \xmark & 69.2 & 40.0 \\
CALIPSO \cite{calipso} & \xmark & - & 46.4 \\
PMFNet \cite{wan2019pose} & \xmark & - & 52.0 \\
ConsNet \cite{liu2020consnet} & \xmark & - & 53.2 \\
MLCNet \cite{sun2020human} & \xmark & - & 55.2 \\
\hline
UnionDet \cite{kim2020uniondet} & \cmark & - & 47.5 \\
DIRV \cite{dirv} w/o prior & \cmark & 64.7 & 54.8 \\
DIRV \cite{dirv} w/o flip & \cmark & 64.1 & 55.2 \\
DIRV \cite{dirv} & \cmark & 64.6 & 56.1 \\
\hline
\textbf{DIABOLO (ours)} & \cmark & \textbf{76.7} & \textbf{57.3} \\
\hline
\end{tabular}
\end{center}
\end{table}

\subsection{Ablation studies for DIABOLO learning strategies}

UnionDet \cite{kim2020uniondet} and DIRV \cite{dirv} methods first train the object detector on COCO dataset and then freeze both backbone and instance detection branch to maintain detection performance. 
In this work, we choose to jointly learn detection and interaction on V-COCO and COCO datasets. To measure the impact of this strategy, we use EfficientDet-D1 backbone since training time is much faster than D3. 
For a fine-grained analysis of results, metrics are also computed with perfect ground truth (GT) detections, to uncorrelate instance and interaction detection performances. 
As shown in table \ref{multi_dataset}, the architecture of DIABOLO interaction branch poorly works with a frozen backbone and detection branch since the $AP_{role}$ is only 46.2\% with perfect detections. However, DIABOLO trained without freezing backbone nor detection branch, achieves 61.3\% with perfect detection but only 43\% with model detections. Such a result is explained by poor instance detection results (18\% of mAP on V-COCO test set) of DIABOLO when learned on V-COCO only for both tasks.
When keeping feeding DIABOLO with COCO along with V-COCO, instance detection performance gets back to normal (35\% of mAP) and $AP_{role}$ reaches 51.1\% with model detections, showing the positive effect of joint learning.

\begin{table*}
\caption{Ablation studies for DIABOLO multi-task learning strategies. Metrics are evaluated on V-COCO test set.}
\label{multi_dataset}
\begin{center}
\begin{tabular}{|c|C{2cm}|C{1.5cm}|C{1cm}|C{1.5cm}|C{2cm}|C{1.5cm}|C{2cm}|}
\hline
\textbf{Backbone} & \textbf{Detection branch} & \textbf{Batch size} & \textbf{mAP (\%)} & \textbf{$AP_{agent}$ (\%)} & \textbf{$AP_{agent}$ with GT detections (\%)} & \textbf{$AP_{role}$ (\%)} & \textbf{$AP_{role}$ with GT detections (\%)}\\
\hline
\hline
EfficientDet-D1 & frozen & 12 V-COCO & 42 & 59.8 & 61.9 & 37.3 & 46.2 \\
EfficientDet-D1 & learned & 12 V-COCO & 18 & 71.1 & 77.5 & 43.0 & 61.3 \\
EfficientDet-D1 & learned & 12 V-COCO + 16 COCO & 35 & 75.7 & 79.5 & \textbf{51.1} & \textbf{64.0} \\
\hline
EfficientDet-D3 & learned & 24 V-COCO + 32 COCO & 47 & \textbf{76.7} & \textbf{80.2} & \textbf{57.3} & \textbf{66.0}\\
\hline
\end{tabular}
\end{center}
\end{table*}



To further assess
the part of error in interaction detection induced by instance detection failure, we train DIABOLO
with a better backbone, EfficientDet-D3, and once again compare results 
given by the model detections with those with perfect detections (cf. Table~\ref{multi_dataset}).
DIABOLO instance detection reaches 47\% of mAP on V-COCO test set. 
Subject-target association task is more heavily impacted by instance detection than verb estimation. Indeed, using ground truth detection only increases by 3.5 p.p $AP_{agent}$ whereas improvement is 8.7 p.p. for $AP_{role}$. Additionally, interaction branch with a perfect detection reaches 80.2\% for $AP_{agent}$ but only 66.0\% for $AP_{role}$, which leaves room for improvement on the subject-target association task.

\subsection{Results of new baseline DIABOLO on \hho{} dataset}

\subsubsection{Quantitative results}

We evaluate DIABOLO on the new dataset \hho{}.
DIABOLO achieves $AP_{agent}$ scores of respectively 41\% and 40.6\% for ``Original" and ``Objectness" scenarios. For $AP_{role}$ metric, the scores are respectively 25.26\% and 23.68\% for ``Original" and ``Objectness" scenarios.
The overall $AP_{role}$ is lower than the one on V-COCO, suggesting that
\hho{} is more challenging than V-COCO. 
We think this is due to \hho{} annotation which is more exhaustive (number of interactions per image has more than doubled), and taxonomy which is more related to the body attitude of the subject, rather than the interaction context which is no longer an extra clue.
Detailed performance per verb will be made available along with \hho{} dataset.

\subsubsection{Qualitative results}

\begin{figure}[h!]
\centering
\includegraphics[width=8.5cm]{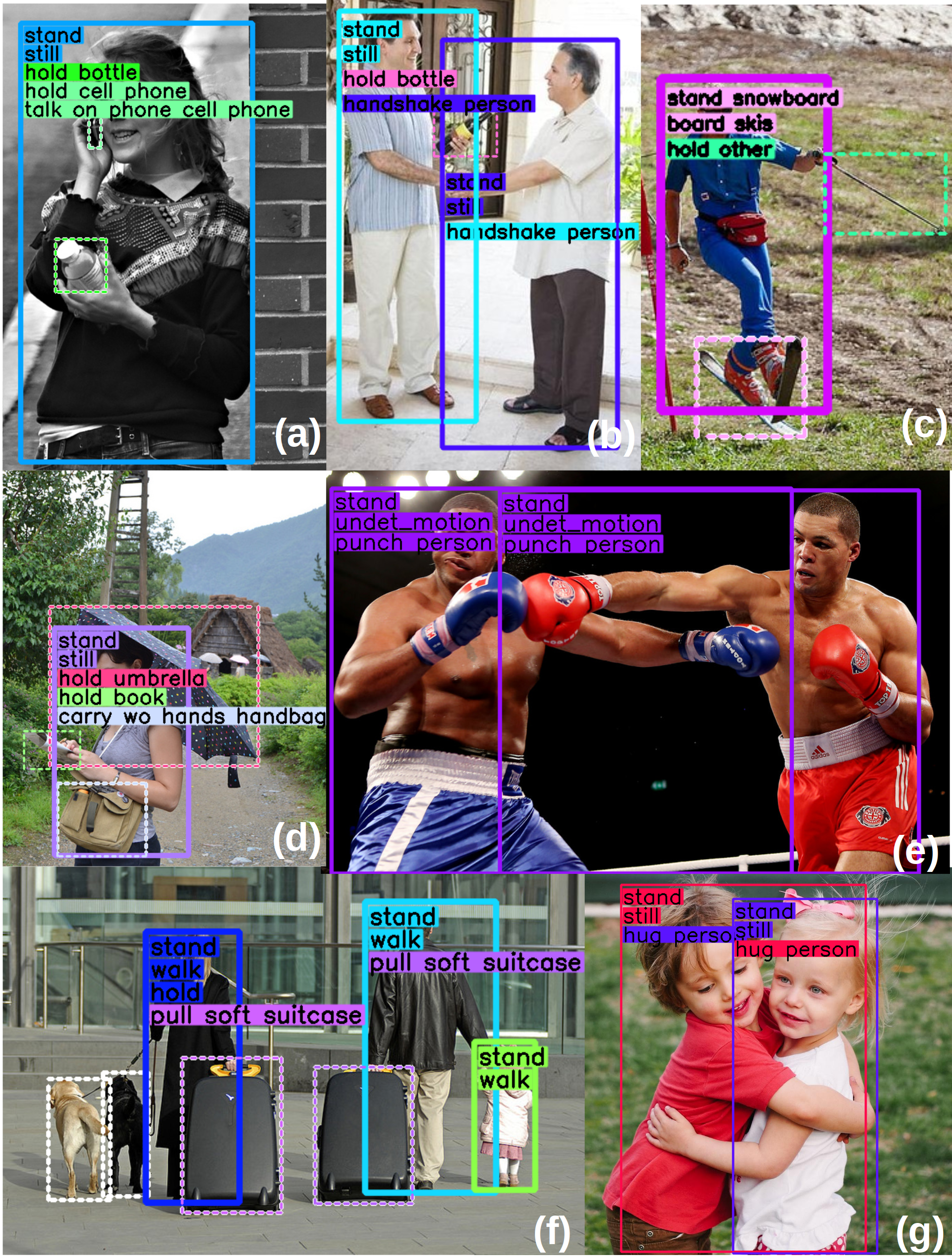}
\caption{Qualitative results of DIABOLO on \hho{}. Interaction name has the color of the target bounding box if any, otherwise the color of the subject box. Dashed-(solid-)line box is for object (human, resp.). White boxes correspond to detected objects not interacting. }
\label{diabolo_on_h2o}
\end{figure}

Figure \ref{diabolo_on_h2o} shows qualitative results of DIABOLO learned on \hho{}. In sample (b), (e) and (g), we can see that DIABOLO well detects \hhi{} and their reciprocity as ``hug'', ``punch'', ``highfive'' and ``handshake''. In illustration (a), DIABOLO is able to detect that two objects are held. Similarly, in example (d), both book and umbrella are correctly detected as held. In sample (c), \hho{} taxonomy allows DIABOLO to correctly detect a person standing on his/her skis. 
Finally, in example (f), DIABOLO correctly predicts that the suitcases are pulled and not held. This illustrates the fact that \hho{} taxonomy is 
closer to the body attitude and the way of interacting with a target than the interaction context.

\subsection{Computation time}

Table \ref{time} shows computation times of our method compared to DIRV \cite{dirv}. 
Both methods are run on a NVIDIA RTX2080Ti GPU. 
DIABOLO inference time with EfficientDet-D3 is competitive with DIRV. 
To obtain the best performance \cite{dirv} apply their method on the image and its flipped version (cf. Table~\ref{results_vcoco}). With such a strategy, measured inference time for DIRV method is 132ms, and 129ms for DIABOLO.

\begin{table}[h!]
\caption{Computation time}
\label{time}
\begin{center}
\begin{tabular}{|l|c|c|c|}
\hline
\textbf{Method} & \textbf{Backbone} & \textbf{Inference Time (ms)}\\
\hline
\hline
DIRV \cite{dirv} w/o flip & EfficientDet-D3 & 108\\
DIRV \cite{dirv} & EfficientDet-D3 & 132 \\
\hline
DIABOLO & EfficientDet-D1 & 89\\
DIABOLO & EfficientDet-D3 & 129\\
\hline
\end{tabular}
\end{center}
\end{table}

\section{CONCLUSION}

In this paper, we propose a new challenging dataset called \hho{} that copes at the same time with human interactions with other people and objects.
All these interactions follow a novel taxonomy focusing on the subject's body attitude rather than the type of the object involved or the environment. 
As a first baseline for \hho{}, we propose DIABOLO, a new multi-task method to detect both instances and interactions without needing an external detector.
This is a single-shot subject-centric method running in a fast and constant time independently of the number of instances in the image.
Finally, we experimentally show that jointly training interaction and instance detections largely improves interaction detection, making DIABOLO a strong approach outperforming the state of the art on V-COCO dataset.

\section{ACKNOWLEDGMENTS}

This publication was made possible by the use of the FactoryIA supercomputer, financially supported by the Ile-de-France Regional Council. Moreover, this work benefited from a government grant managed by the French National Research Agency under the future investment program with the reference ANR-19-STHP-0006.


{\small
\bibliographystyle{ieee}
\bibliography{bib}
}

\end{document}